\documentclass{article}

\usepackage{arxiv}

\usepackage[utf8]{inputenc} 
\usepackage[T1]{fontenc}    
\usepackage{hyperref}       
\usepackage{url}            
\usepackage{booktabs}       
\usepackage{bm} 
\usepackage{bbm}
\usepackage{pifont}
\usepackage{mathtools}
\usepackage{siunitx}
\usepackage{amsmath,amsfonts}
\usepackage{nicefrac}       
\usepackage{microtype}      
\usepackage{cleveref}       
\usepackage{graphicx}
\usepackage{doi}
\usepackage{tabularx}
\usepackage{multirow}
\usepackage{lmodern}
\usepackage{dirtytalk}
\usepackage{xcolor}
\usepackage{enumitem}
\usepackage{float}

\newcommand{\cmark}{\ding{51}}%
\newcommand{\xmark}{\ding{55}}%
\title{The Deep Promotion Time Cure Model}

\date{}

\author{ \href{https://orcid.org/0000-0002-3695-1436}{\includegraphics[scale=0.06]{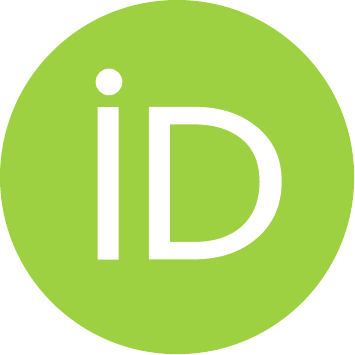}\hspace{1mm}Victor Medina-Olivares} 
	\\
	Chair of Uncertainty Quantification and Statistical Learning\\
	Research Center Trustworthy Data Science and Security (UA Ruhr)\\
	Department of Statistics (TU Dortmund)\\
	Joseph-von-Fraunhofer-Str. 25\\
	44227 Dortmund, Germany \\
	\texttt{victor.medina@tu-dortmund.de} \\
	\And
	\href{https://orcid.org/0000-0001-7685-262X}{\includegraphics[scale=0.06]{orcid.pdf}\hspace{1mm}Stefan Lessmann} \\
	Chair of Information Systems\\
	Humboldt-Universität zu Berlin\\
	Unter den Linden 6\\
	10099 Berlin, Germany \\
	\AND
	\href{https://orcid.org/0000-0002-5072-5347}{\includegraphics[scale=0.06]{orcid.pdf}\hspace{1mm}Nadja Klein}\\
	Chair of Uncertainty Quantification and Statistical Learning\\
	Research Center Trustworthy Data Science and Security (UA Ruhr)\\
	Department of Statistics (TU Dortmund)\\
	Joseph-von-Fraunhofer-Str. 25\\
	44227 Dortmund, Germany \\
}

\renewcommand{\shorttitle}{Deep-PTCM}

\hypersetup{
pdftitle={The Deep Promotion Time Cure Model},
pdfsubject={q-bio.NC, q-bio.QM},
pdfauthor={Victor Medina-Olivares, Stefan Lessmann, Nadja Klein},
pdfkeywords={credit risk, cure models, deep learning, interpretability, survival analysis},
}

\begin{document}
\maketitle

\begin{abstract}
We propose a novel method for predicting time-to-event in the presence of cure fractions based on flexible survivals models integrated into a deep neural network framework. Our approach allows for non-linear relationships and high-dimensional interactions between covariates and survival and is suitable for large-scale applications. Furthermore, we allow the method to incorporate an identified predictor formed of an additive decomposition of interpretable linear and non-linear effects and add an orthogonalization layer to capture potential higher dimensional interactions. We demonstrate the usefulness and computational efficiency of our method via simulations and apply it to a large portfolio of US mortgage loans. Here, we find not only a better predictive performance of our framework but also a more realistic picture of covariate effects.%
\end{abstract}


\section{Introduction}
Lenders employ mathematical models to assist decision-making by estimating each customer's probability of a credit event. These models, known as credit scoring systems, were initially developed to predict the probability of default for specific products. 

Over time, the use and purpose of these systems have become more diverse and aligned with the lender's strategic goals. Moreover, new computational advancements and the pursuit of better models have urged research on deep learning (DL) approaches in this field. Gunnarsson et al.~\cite{Gunnarsson_ejor_21}, conducting a study comparing DL algorithms and their practicality in credit scoring, find that while ensemble methods are still favored, DL approaches have potential, e.g., in handling less traditional data sources. Meanwhile, Stevenson et al.~\cite{Stevenson_ejor_21} reveal the merit of DL for predicting default in small businesses using text data. In another study by Korangi et al.~\cite{Korangi_ejor_23}, transformer models were employed to process time-varying covariates, such as accounting metrics from the balance sheet, to predict the bankruptcy of middle-capitalization companies, showing better performance than traditional models. These findings are part of the recent evidence suggesting that DL techniques are promising to improve credit scoring systems and expand the range of data types that can be leveraged in this field.

Until now, most applied DL models to credit risk have focused on classification tasks, where a pre-defined performance period of a binary decision is established. A different route is that of survival analysis for building scoring systems but is less explored in the DL context \cite{blumenstock2022deep}. Here, the outcome of interest is the time until an event occurs. One challenge in survival analysis is to reliably describe the distribution of survival times, trying to convey, for example, if all subjects are prone to the event of interest. In credit risk modeling, it is however natural to expect that some borrowers will never experience the event, resulting in heavy censoring at the end of the study \cite{dirick_macro-economic_2019}. In this situation, cure rate models are preferred \cite{farewell_use_1982}, which extend survival models by including a latent cure fraction. The advantage is that these models allow to separate the factors that influence the probability of the event occurrence from those that affect its timing.

Another challenge is understanding how the subject-specific features (or covariate effects) relate to survival times. To this end, two main classes of cure models exist: the mixture cure model (MCM) \cite{boag_maximum_1949} and the promotion time cure model (PTCM) \cite{yakovlev_stochastic_1996,tsodikov2001estimation}. Although the MCM has been extensively studied in the credit risk literature (see Table \ref{tab:refs_cure}), the PTCM, introduced in the late 1990s, has gone practically under the radar and is the focus of this paper.

The MCM assumes a binary response variable in the population that describes those \textit{cured} and those susceptible to the event. This approach has been broadly developed in parametric, semi-parametric, and non-parametric formulations, and to handle continuous, discrete, and longitudinal data (see \cite{amico_cure_2018} for a comprehensive review). In contrast, the PTCM, which originates from cancer studies, assumes each subject has unobserved competing risk factors, such as cancer cells. In this situation, a cured patient will have zero cancer cells, while a susceptible patient's event will occur when the first cell develops into a palpable cancer mass. Although initially conceived for tumors, its statistical principles apply to broader contexts. For example, in credit-related applications, competing risk factors include causes for borrower default, such as job loss, inability to work, strategic default, and failed businesses \cite{barriga2015non}.

We make three significant contributions in this manuscript: two methodological and one empirical. From a methodological standpoint, first, we reformulate the PTCM using a deep neural network (DNN) architecture. We label our approach \textit{Deep-PTCM}.

The second methodological contribution allows the user to decompose the predictor as linear and non-linear components, with the latter estimated through a DNN. This separation aims to facilitate the interpretation of covariate effects, a common criticism when applying DL approaches. However, it is known that a neural network (NN) can approximate any continuous function \cite{cybenko_approximation_1989}, in particular, a linear one. Hence, to avoid identifiability issues, we follow \cite{rugamer2022semi} and add an orthogonalization layer that projects the non-linear component's output into the linear one's orthogonal complement. The orthogonalization is performed using a QR decomposition, which is stable when computed in a mini-batch training routine \cite{roberts2020qr}.

From an empirical perspective and to the best of our knowledge, this is the first study to apply such a general and flexible framework of PTCMs in the credit risk context. First, most of the cure models studied in credit scoring belong to the class of MCMs, leaving the PTCM relatively unexplored. However, we do not find any solid justification in the literature for choosing MCM over PTCM, and its preference may be due to its popularity. Second, none of the cure models, regardless of the class selected, allow for complex and often more realistic non-linear relationships and interactions between covariates and survival. As we show later, this assumption limits the predictive power, an essential aspect of credit risk management \cite{thomas2017credit}. Concretely, we build a cure model to predict the time to default in a large US mortgage portfolio. We show that the Deep-PTCM significantly outperforms the standard PTCM in calibration and discrimination. 

Overall, our {Deep-PTCM}, has the following highly relevant advantages over existing competitors: 

\begin{enumerate}[label=(\roman*)]
	\item It generalizes the standard PTCM, which assumes linear dependency in its predictor. This can be seen as a one-layer NN with one unit. 
	\item It provides more flexibility than traditional estimation pipelines by replacing data preprocessing and feature engineering with a differentiable loss function estimated via gradient descent. That facilitates the model to incorporate structured and unstructured data such as text and images, increasing the scope of its applicability. 
	\item It is scalable since all model parameters are integrated into an end-to-end DNN, making the estimation procedure computationally efficient and easily parallelizable (GPUs/TPUs). 
\end{enumerate}

This provides comparative advantages over recent efforts from medical research by Xie and Yu \cite{xie_promotion_2021}. The authors propose a PTCM with a DNN component and show it can increase the model's performance compared with non-parametric approaches with splines. However, this is the only work in this respect, indicating that the interface of PTCMs and DNNs is an underexplored area from both a modeling and an application perspective. An example is the estimation procedure, carried out iteratively using the Expectation-Maximization (EM) algorithm introduced by \cite{chen_maximum_2001}, where the DNN is optimized at each maximization step. That results in a computationally inefficient procedure that limits the approach's materiality in increasingly prevalent big data environments. Through a simulation study, we demonstrate that Deep-PTCM scales better than the approach proposed in \cite{xie_promotion_2021}. This improvement permits us to estimate the model on a training set with approximately 150k borrowers, the largest in this context (see Table \ref{tab:refs_cure}), in a few minutes rather than hours. 
\begin{enumerate}[label=(\roman*)]
    \setcounter{enumi}{3}
    \item Its implementation uses the TensorFlow framework \cite{tensorflow2015-whitepaper}, making it easy to accommodate all layers, optimizers, and features available there.
\end{enumerate}

The rest of the paper is organized as follows. In Section \ref{sec:meth}, we describe the PTCM, its reformulation in an end-to-end DNN framework, and how we estimate it efficiently even when the dataset is large. Section \ref{sec:sim} offers two simulation studies, one comparing our estimation approach to Xie and Yu's \cite{xie_promotion_2021}, and the second studies how the Deep-PTCM can recover the linear effects when the orthogonalization step is included. In Section \ref{sec:app}, we present our credit risk study, while Section \ref{sec:concl} concludes.

\section{Related work}
Although the contributions presented have the potential for applications beyond the credit-related context, the motivation for this work arises from the importance of credit scoring models in a predominantly data-driven industry and the lack of studies combining cure models and DL.

Most of the cure models applied so far belong to the class of MCMs. Tong et al.~\cite{tong_mixture_2012} introduce the MCM and compare its performance to the logistic regression and the Cox Proportional Hazard model (Cox PH), noting the ability to distinguish among borrower's susceptibility is appealing for risk management. Similarly, Dirick et al.~\cite{dirick_time_2017} compare the performance of different survival approaches in ten datasets. They find comparable performance between the MCM, Cox PH, and Accelerated Failure Time models, with a promising economic performance by the MCM.

Moreover, Louzada et al.~\cite{louzada_modeling_2014} demonstrate that the flexibility of the MCM allows modeling survival data even when the proportional hazard assumption is not satisfied. Extensions to include exogenous time-varying covariates can be found in \cite{de_leonardis_default_2014}, from a discrete-time perspective, and in \cite{dirick_macro-economic_2019}, for the continuous-time one. Furthermore, Zhang et al.~\cite{zhang_new_2019} introduce a new MCM to allow the non-cured borrowers to be susceptible to a subset of risks instead of all of them as it is commonly assumed in competing risk settings. 

Although much of the literature focuses on the MCM, some work, all from the same group, have studied the PTCM. Namely, in \cite{oliveira2014evidence}, a PTCM is applied to relate the intensity of default and recovery rates in a Brazilian loan portfolio. This study, however, does not include covariates in the model. In addition, in \cite{barriga2015non}, different activation mechanisms of the PTCM are analyzed. A bivariate survival process is considered in \cite{cancho2016non}, and Ribeiro de Oliveira Jr et al.~\cite{ribeiro2017zero} extend it to account for events in time zero. More recently, Toledo et al.~\cite{toledo2022gompertz} allow for early events with fractions incorporating covariates effects. 

Table \ref{tab:refs_cure} compares the various approaches involving cure models with application to credit risk. All these contributions, whether with MCM or PTCM, consider covariate linear effects, an assumption that is often unrealistic and too simplistic. Furthermore, regarding cure rate models that include non-linear effects, traditional non-parametric methods, such as smoothing splines, do not scale well with high-dimensional interactions \cite{xie_promotion_2021}. In this regard, the Deep-
PTCM can not only capture complex covariate effects but is also scalable by leveraging the flexibility of DL.

\begin{table*}[ht!]
\caption{List of references in the credit risk literature with cure models. T|N|X refer to the maximum performance periods (months), the sample size, and the number of covariates (before preprocessing).}
\label{tab:refs_cure}
\centering
\begin{tabularx}{\textwidth}{XcXXXcc}
\toprule
\textbf{Reference} & \textbf{Class} & \textbf{Data} & \textbf{T|N|X} & \textbf{Metric(s)} & \textbf{Non-linear}& \textbf{DL} \\ \midrule
Tong et al.~\cite{tong_mixture_2012} &  MCM & UK Personal loans & 36|27527|14 & AUC; H-measure; KS & \xmark & \xmark \\
De Leonardis et al.~\cite{de_leonardis_default_2014} &  MCM & SMEs & 84|27579|9 & AUC & \xmark & \xmark  \\ 
Louzada et al.~\cite{louzada_modeling_2014} &  MCM & Brazilian Personal loans & 12|40115|2 & Expected loss & \xmark & \xmark  \\ 
Liu et al.~\cite{liu_identifying_2015} &  MCM & Chinese Mortgage loans & 36|14068|14 & Gini; KS & \xmark & \xmark   \\ 
Dirick et al.~\cite{dirick_akaike_2015} &  MCM & UK Personal loans & 36|7521|8 & AIC; AUC & \xmark & \xmark   \\ 
Wycinka \& Jurkiewicz \cite{wycinka_mixture_2017} &  MCM & Polish Personal loans & 24|5000|12 & AUC; H-measure & \xmark & \xmark  \\ 
Dirick et al.~\cite{dirick_time_2017} &  MCM & 10 datasets (personal loans and SMEs) &  60|80641|31 & AUC; MAE; MSE; Financial metrics  & \xmark & \xmark \\ 
Dirick et al.~\cite{dirick_macro-economic_2019} &  MCM & Belgian Personal loans &  36|20000|13 & AIC & \xmark & \xmark  \\ 
Zhang et al.~\cite{zhang_new_2019} &  MCM & P2P loans &  36|50000|5 & AIC  & \xmark & \xmark  \\ 
Jiang et al.~\cite{jiang_prediction-driven_2019} &  MCM & P2P loans &  12|52573|31 & AUC; Concordance corr.; H-measure; KS  & \xmark & \xmark  \\ 
Dirick et al.~\cite{dirick_hierarchical_2022} &  MCM & UK Personal loans &  36|7521|8 & AIC; BIC; MAE; MSE & \xmark & \xmark  \\
Oliveira \& Louzada \cite{oliveira2014evidence} &  PTCM & Brazilian Personal loans &  84|20000|0 & Expected LGD & \xmark & \xmark  \\
Barriga et al.~\cite{barriga2015non} &  PTCM & Brazilian Personal loans &  36|236|1 & AIC; BIC; RMSE & \xmark & \xmark  \\
Cancho et al.~\cite{cancho2016non} &  PTCM & Brazilian customer data &  36|1188|2 & AIC; BIC; DIC & \xmark & \xmark \\
Ribeiro de Oliveira Jr et al.~\cite{ribeiro2017zero} &  PTCM & Brazilian Personal loans &  60|5733|3 & AIC & \xmark & \xmark \\
Toledo et al.~\cite{toledo2022gompertz} &  PTCM & Brazilian loans &  24|9645|4 & AIC; BIC & \xmark & \xmark \\
\textbf{This work} &  PTCM & US Mortgage loans &  \textbf{154}|\textbf{149561}|19 & $\text{AUC}_{cure}$; IBS & \cmark & \cmark  \\
\bottomrule

\end{tabularx}
\end{table*}

\section{Methodology} \label{sec:meth}

\subsection{The Promotion Time Cure Model} \label{subsec:ptcm}
The PTCM assumes that a subject has $K$ unobserved competing risk factors, each of which can lead to the event's occurrence. Furthermore, suppose that $K$ is distributed as a Poisson with mean $\theta$ and denote by $Y_{k}$, $k=1,\ldots, K$ the random time for the $k$th risk factor. Given $K$, it is assumed that the random variables $Y_{k}$ are independently distributed with cumulative distribution function (CDF) $F(t)$. The time to event $T^*$ is the time elapsed until the first unobserved competing risk factor is triggered, i.e.\ $T^*=\min\{Y_{0}, Y_{1},\ldots, Y_{K}\}$, where $P(Y_{0}=\infty) = 1$. The survival function, $S_p(t)=P(T^*>t)$, is represented by
\begin{equation} \label{eq:1}
  \begin{aligned}
  S_p(t) &= P(\text{no event by time } t)\\
  &= P(K=0)+P(Y_1>t,...,Y_K>t, K\ge 1) \\
  &=\exp(-\theta)+\sum_{j=1}^{\infty} \frac{[(1-F(t))\theta]^j}{j!}\exp(-\theta)\\
  &= \exp(-\theta F(t)),
  \end{aligned}
\end{equation}
and the cure fraction is $\lim\limits_{t \to \infty} S_p(t)=\exp(-\theta)$. Note that, since $\lim\limits_{t \to \infty} S_p(t)$ can be positive, $S_p(\cdot)$ is not a proper survival function. We deliberately call it the \textit{survival function of the population} and add the subindex $p$ to differentiate it from $S(t)=1-F(t)$, the (proper) survival function of the risk factors.

Denote by $\mathbf{x}$ the $q$-dimensional vector of covariates. Then the PTCM in its standard version relates $\theta$ with $\mathbf{x}$ through $\theta(\mathbf{x}) = \exp(\mathbf{w}^\top \mathbf{x}+b)$, where $\mathbf{w}\in \mathbbm{R}^q$ is the vector of regression coefficients and $b \in \mathbbm{R}$ the intercept \cite{chen_new_1999}. Note that the hazard function of the population is then $h_p(t;\mathbf{x}) = \exp(\mathbf{w}^\top \mathbf{x}+b)f(t)$, with $f(t)=\frac{\mathrm{d}F(t)}{\mathrm{d}t}$. Since the hazard function preserves proportionality, the PTCM is also known as the proportional hazard cure model \cite{peng_cure_2021}.

The PTCM has been studied and extended in several directions in the statistical community. For example, an EM algorithm to estimate the model with missing covariates \cite{chen_maximum_2001}, extensions to handle interval-censored data \cite{liu_semiparametric_2009} and to include random effects \cite{carvalho_lopes_random_2012}, or categorical time-varying covariates \cite{lambert_inclusion_2020}, latent risk classes \cite{kim_new_2009}, and longitudinal covariates \cite{brown_bayesian_2003} have been developed. Moreover, non-parametric approaches have recently been explored, in particular, to model univariate covariate effects by including smoothing splines \cite{chen_promotion_2018, bremhorst_nonparametric_2019}. Still, these do not scale well in large-scale applications.

\subsection{The Deep-PTCM} \label{subsec:deepptcm}
It is commonly assumed that there is a linear relationship between covariates and survival, which facilitates interpretation. However, this also restricts the regression predictor to a specific structure. To address this limitation, we redefine the PTCM as an end-to-end DNN architecture. This approach lets us consider non-linear relationships and high-dimensional interactions between covariates and survival. Furthermore, we develop an efficient estimation algorithm using existing DL libraries, allowing the framework to be applied to large and unstructured datasets.

Consider that subject $i$ in the population ($i=1,\ldots,N$) has a vector of covariates $\mathbf{x}_i \in \mathbbm{R}^q$, and a number of unobserved competing risk factors $K_i \sim \text{Pois}(\theta(\mathbf{x}_i)), \theta(\mathbf{x}_i)>0$. Moreover, denote as $Y_{ik}$, ($k = 1,\ldots,K_i$), the associated random event time for the $k$th risk factor with CDF $F(t)$, $\forall k$. As in the PTCM, the uncensored time to event for subject $i$ is $T^*_{i} = \min\{Y_{i0}, Y_{i1},\ldots, Y_{iK_i}\}$, with $P(Y_{i0}=\infty) = 1$. Assuming a right-censoring mechanism and denoting by $C_i$ the censored time for subject $i$, the observational event time follows $T_i=\min\{C_i, T^*_{i}\}$. Denote the realization of $T_i$ by $t_i$ and by $\delta_i$ the event indicator that is 1 if the event occurs at time $t_i$ and 0 otherwise. 

The \textit{survival function of the population} follows analogously to Equation \eqref{eq:1}, i.e.\ $S_p(t;\mathbf{x}_i)=\exp(-\theta(\mathbf{x}_i)F(t))$, and the corresponding hazard function is $h_p(t;\mathbf{x}_i) = \theta(\mathbf{x}_i)f(t)$. Yet, we now consider $\theta(\mathbf{x}_i)=\exp(\eta(\mathbf{x}_i))$, with $\eta:\mathbbm{R}^{q}\rightarrow \mathbbm{R}$ a general continuous function rather than a linear combination of the covariates $\mathbf{x}_i$. 

We proceed similarly to \cite{xie_promotion_2021} and model $\eta$ through a DNN. From the universal approximation theorem \cite{cybenko_approximation_1989}, we know that a NN, under certain conditions, can approximate any continuous function. Moreover, the linear specification described in Section \ref{subsec:ptcm}, $\eta(\mathbf{x}_i) = \mathbf{w}^\top\mathbf{x}_i+b$, can also be parameterized with a NN in which only one layer and one neuron are needed. Therefore, the Deep-PTCM can be regarded as a generalization of the standard PTCM. 

The Deep-PTCM is flexible enough to represent complex covariate relationships and/or handle unstructured data. However, there are situations in which the interest is to identify if the predictor $\eta$ has structured linear effects to ease interpretation. To this aim, we also allow the framework to estimate $\eta$ as an identifiable sum between linear ($\eta^{lin}$) and non-linear ($\eta^{non}$) predictors. 

To avoid potential identifiability issues specifically, we follow the orthogonalization procedure in \cite{rugamer2022semi}. This is accomplished by calculating orthogonal projection matrices $\mathcal{P},\mathcal{P}^\perp \in \mathbbm{R}^{N\times N}$, such that $\bm{\eta}^{lin} = \mathcal{P}\bm{\eta}$ and $\bm{\eta}^{non} = \mathcal{P}^\perp \bm{\eta}$, where $\bm{\eta} = (\eta(\mathbf{x}_1),\ldots,\eta(\mathbf{x}_N))^\top$ ($\bm{\eta}^{lin}$ and $\bm{\eta}^{non}$ follow analogously). The projection matrices are obtained by performing a QR decomposition of 
$\tilde{\mathbf{X}}=[\mathbbm{1}_N,\mathbf{X}]=[\mathbbm{1}_N,(\mathbf{x}_1,\ldots,\mathbf{x}_N)^\top]$, i.e.\ $\tilde{\mathbf{X}}=QR$, where $Q$ and $R$ are orthonormal and upper-triangular matrices, respectively, and $\mathbbm{1}_N  \in \mathbbm{R}^{N}$ is a vector of ones (intercept). Hence, we obtain the projection matrices as $\mathcal{P}=QQ^\top$ and $\mathcal{P}^{\perp}=I-QQ^\top$. While the QR decomposition is usually performed on the entire set $\mathbf{X}$, it has been shown that this decomposition can be computed stably in a mini-batch training routine \cite{rugamer2022semi,roberts2020qr}, and thus can be readily included in the learning procedure.

\begin{figure}[ht]
    \centering
    \includegraphics[width=0.7\textwidth]{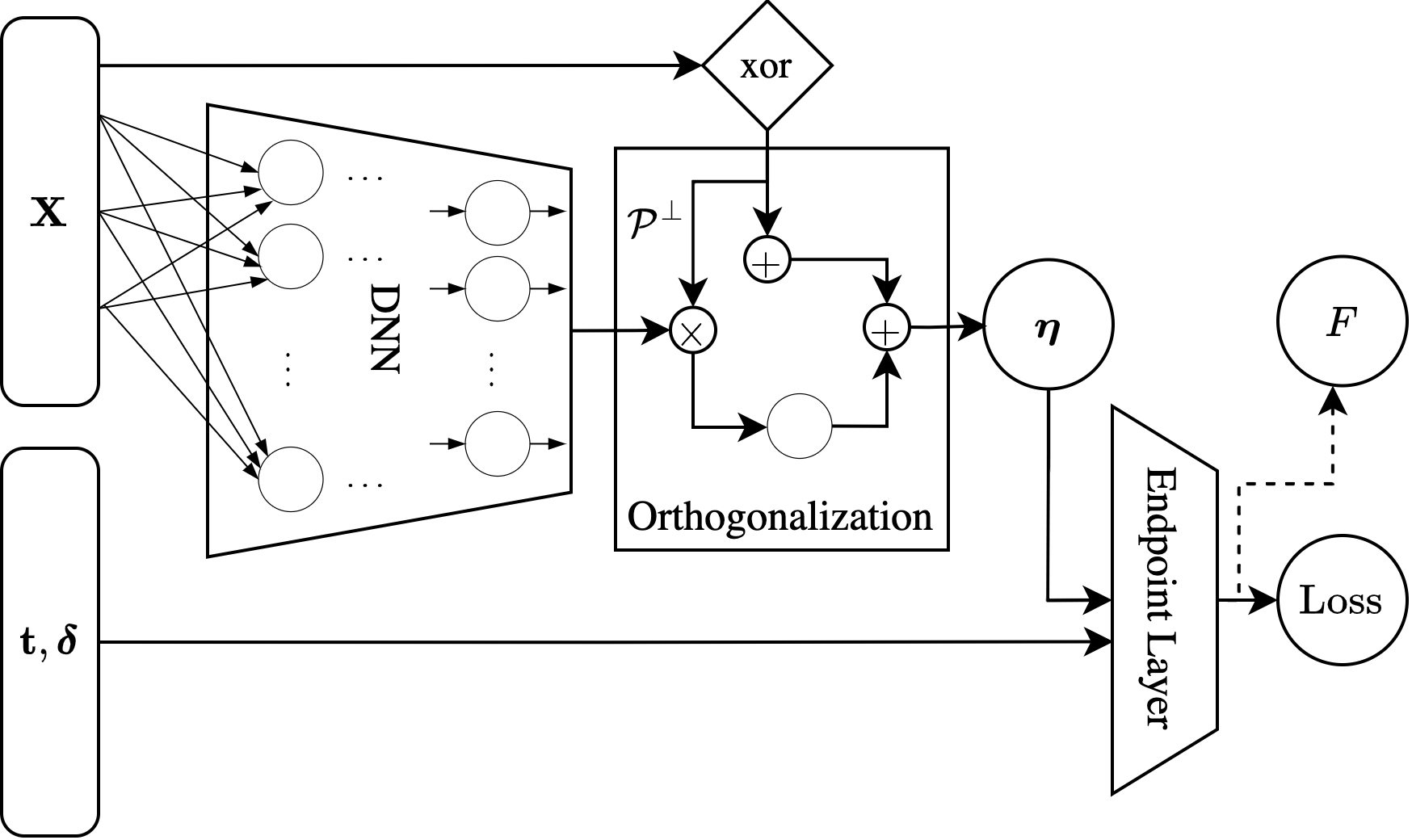}
    \caption{Generic representation of the Deep-PTCM architecture.}
    \label{fig:nnet}
\end{figure}

Figure \ref{fig:nnet} depicts the main idea of the Deep-PTCM architecture. The \texttt{DNN} block has as inputs the covariates $\mathbf{X}$ and is here where we define the appropriate architecture for the data at hand. For example, if we are provided with unstructured data, such as images, we may include in this block convolutional NNs \cite{goodfellow2016deep}. The output of this block goes to the \texttt{Orthogonalization} layer. If the orthogonalization step is required, then $\bm\eta$ is built by the sum of a linear predictor and the appropriate projection of the \texttt{DNN} block output into the orthogonal complement of that linear predictor (e.g., a subset of covariates). If not, the whole predictor $\bm\eta$ is estimated without decomposition ($\mathcal{P}^{\perp}=I$). 

For illustrative purposes, let us consider a fully connected feedforward network (FCFN) with $L$ hidden layers. Specifically, suppose that layer $l$ ($l=1,\ldots,L)$ has $n_l$ neurons, hence the $l$th layer, $\mathbf{g}^{(l)}:\mathbbm{R}^{n_{l-1}}\rightarrow \mathbbm{R}^{n_{l}}$ ($n_0=q$), follows
\begin{displaymath}
    \mathbf{g}^{(l)}(\mathbf{z})=\left[g_1^{(l)}(\mathbf{z}),\ldots,g_{n_l}^{(l)}(\mathbf{z})\right]^{\top},
\end{displaymath}
with $g_m^{(l)}(\mathbf{z})=a^{(l)}(\mathbf{w}_m^{{(l)}\top}\mathbf{z}+b_m^{(l)})$, $m=1,\ldots,n_l$. Where $a^{(l)}:\mathbbm{R}\rightarrow\mathbbm{R}$ is the activation function for the $l$th layer, $\mathbf{w}_m^{(l)} \in \mathbbm{R}^{n_{l-1}}$ the weights associated to the $m$th neuron of the $l$th layer, and $b_m^{(l)} \in \mathbbm{R}$ the corresponding intercept. Many activation functions have been proposed (see \cite{zhang2021dive}, Chap.~5.1.2). One popular choice is the rectified linear unit (ReLU) \cite{nair2010rectified}, which is the non-linear transformation defined by $a(x)=\max(0,x)$.

Therefore, when a FCFN is considered in the \texttt{DNN} block, its output is a vector $\tilde{\bm\eta} \in \mathbbm{R}^{n_L}$ formed by the composition of the $L$ hidden layers, i.e.\ $\tilde{\bm\eta}(\mathbf{x}_i)=\left(\mathbf{g}^{(L)}\circ\cdots\circ\mathbf{g}^{(2)}\circ \mathbf{g}^{(1)}\right)(\mathbf{x}_i)$. Moreover, if no orthogonalization is performed, $\eta(\mathbf{x}_i)$ is finally computed through a final unit with a linear activation, i.e.\ $\eta(\mathbf{x}_i)=\mathbf{w}^{{(L+1)}\top}\tilde{\bm\eta}(\mathbf{x}_i)+b^{(L+1)}$. On the other hand, if orthogonalization is carried out for all covariates, then 
\begin{displaymath}
\bm\eta=\mathbf{X}\mathbf{w}^{lin}+b^{lin}\mathbbm{1}_N+\mathcal{P}^{\perp}\begin{pmatrix}
        \tilde{\bm\eta}(\mathbf{x}_1)^\top\\
        \vdots\\
        \tilde{\bm\eta}(\mathbf{x}_N)^\top
    \end{pmatrix} \mathbf{w}^{(L+1)},
\end{displaymath}
where $\mathbf{w}^{lin}\in \mathbbm{R}^{q}$ and $b^{lin}\in \mathbbm{R}$ are, respectively, the vector of linear coefficients and the intercept.

Moreover, the \texttt{Endpoint Layer} defines the loss function by receiving the inputs $\mathbf{t}=(t_1,\ldots,t_N)^\top$ and $\mathbf{\delta}=(\delta_1, \ldots,\delta_N)^\top$, in addition to $\bm{\eta}$. The loss function for the Deep-PTCM is the negative log-likelihood, introduced below in Equation \eqref{eq:loglik}. Thus, it is in the \texttt{Endpoint Layer} where $F$ is specified in order to calculate the loss. Standard specifications in the PTCM context are the Weibull or the piecewise exponential function. Following \cite{xie_promotion_2021,chen_maximum_2001}, we use the latter as described in Section \ref{subsec:est_deep}, but other specifications can be easily accommodated.

\subsection{Estimation of the Deep-PTCM} \label{subsec:est_deep}
Traditionally, estimation in the PTCM is carried out using the EM algorithm, where $K_i$, the number of risk factors for subject $i$, is treated as
missing data \cite{chen_maximum_2001}. As we illustrate in Section \ref{sec:sim}, however, this approach does not scale well when considering NNs and large datasets. To overcome this computational limitation, we present an end-to-end framework to estimate both the predictor $\bm{\eta}$ and the parameters associated with $F$ through the NN optimization problem. 

For that, first, note that the log-likelihood is
\begin{equation}\label{eq:loglik}
    \begin{split}
    l(\bm{\eta}, F) &= \sum_{i=1}^N\delta_i\log(h_p(t_i;\mathbf{x}_i))+\log(S_p(t_i;\mathbf{x}_i))\\
    &=\sum_{i=1}^N\delta_i[\eta(\mathbf{x}_i) + \log(f(t_i))]-\exp(\eta(\mathbf{x}_i))F(t_i).
    \end{split}
\end{equation}
Considering $F$ as a piecewise exponential function, we partition the length of the study in $J$ intervals according to the distribution of the events, i.e.\ $u_0=0<u_1<\ldots<u_J$ with $u_J>\max_{i \in \{1,\ldots,N\} }t_i$. In each interval $(u_{j-1},u_j]$, the hazard function of the competing risk factors is assumed to be constant. Denote these constants by $\lambda_j$,  $j=1,\ldots,J$. Thus, for $t \in (u_{j-1},u_j]$, $F$ and $f$ can be expressed as $F(t) = 1-\exp\left[-\lambda_j(t-u_{j-1})-\sum_{s=1}^{j-1}\lambda_s(u_s-u_{s-1})\right]$, $f(t) = \lambda_j\exp\left[-\lambda_j(t-u_{j-1})-\sum_{s=1}^{j-1}\lambda_s(u_s-u_{s-1})\right]$.

We train this model efficiently using backpropagation \cite{goodfellow2016deep}. This involves (i) initializing the weights of the DNN units randomly, (ii) feeding the input data through the DNN and calculating the loss, (iii) adjusting the weights of the units to minimize the loss in Equation \eqref{eq:loglik}, and (iv) repeating the steps (ii), (iii) of feeding the input data through the DNN, calculating the loss, and adjusting the weights until the loss in a validation set is minimized. 

Once the DNN is optimized, we can use its estimated weights for prediction. In particular, we can retrieve the \texttt{DNN} block to infer the predictor $\bm{\eta}$ for new data. Analogously, we can create any quantity of interest, such as $S_p$ and $S$, by recovering the corresponding parameters of each block. We created a Python package, \texttt{deepcure}, for the estimation of the Deep-PTCM, which is available on \href{https://github.com/vhmedina/deepcure}{GitHub}. The implementation uses TensorFlow, allowing for seamless integration of all available optimizers and additional features provided by the framework.%

\section{Performance Metrics} \label{sec:perf} 
In the empirical study presented in Section \ref{sec:app}, we evaluate the performance of the models under two metrics. The area under the receiver operating curve (AUC) for cure proportions, which measures how well the model distinguishes between cured and non-cured subjects, and the integrated Brier score (IBS), which measures the calibration throughout the whole study period. We describe these metrics in the following.%

\subsection{AUC for Cure Proportions (\texorpdfstring{$\text{AUC}_{\text{cure}}$}{AUC})}
The AUC \cite{fawcett_introduction_2006} is commonly used in survival analysis to evaluate the performance of a corresponding model. However, the classical formulation does not take cure proportions into account. The receiver operating curve can be regarded as the curve formed by the true positive rate (TPR) and the false positive rate (FPR) for all cut-off points $c$ in $[0,1]$. Asano et al.~\cite{asano_assessing_2014} propose the imputation-based AUC for mixture cure models, and \cite{xie_promotion_2021} extend it to the PTCM. This version of the AUC evaluates the TPR and FPR concerning the probability of being cured. Denote the estimated long-term survival probability as $\hat{\pi}(\mathbf{x}_i)\coloneqq \lim\limits_{t \to \infty} \hat{S}_p(t;\mathbf{x}_i)=\exp(-\exp(\hat{\eta}(\mathbf{x}_i)))$, where $\hat\eta(\cdot)$ is the point estimate of $\eta(\cdot)$. Therefore, the estimates of TPR and FPR, for a given cut-off point $c$ are given by
\begin{equation*}
    \begin{split}
        \widehat{TPR}(c) &= \frac{\sum_{i=1}^N \mathbbm{1}(\hat{\pi}(\mathbf{x}_i)\le c)\cdot(1-\hat{\pi}(\mathbf{x}_i) ) }{\sum_{i=1}^N(1-\hat{\pi}(\mathbf{x}_i))}\\
        \widehat{FPR}(c)  &= \frac{\sum_{i=1}^N \mathbbm{1}(\hat{\pi}(\mathbf{x}_i)\le c)\cdot \hat{\pi}(\mathbf{x}_i)  }{\sum_{i=1}^N \hat{\pi}(\mathbf{x}_i)},
    \end{split}
\end{equation*}
where $\mathbbm{1}(\cdot)$ denotes the indicator function. $\text{AUC}_{\text{cure}}$ is calculated using trapezoidal integration over $c \in [0,1]$. 

\subsection{Integrated Brier Score (IBS)}
The Brier score \cite{brier_verification_1950} corresponds to the mean squared error of the predicted probabilities for binary classification. In the survival context, we can estimate whether a subject survives longer or not at a specific time $t$. Moreover, Graf et al.~\cite{graf_assessment_1999} introduced a generalization of the Brier score to handle censoring. This is the version that we use and is specified as 
\begin{displaymath}
    \widehat{BS}(t) = \frac{1}{N}\sum_{i=1}^N\Bigg[\Bigg.\frac{\hat{S}_p(t;\mathbf{x}_i)^2\mathbbm{1}(t_i\le t,\delta_i=1)}{\hat{G}(t_i)}+\frac{(1-\hat{S}_p(t;\mathbf{x}_i))^2\mathbbm{1}(t_i>t)}{\hat{G}(t)} \Bigg.\Bigg],
\end{displaymath}
where $\hat{G}(\cdot)$ is the Kaplan-Meier estimator of the censoring survival function. By integrating the time-dependent Brier score over time, we obtain the Integrated Brier score (IBS) \cite{graf_assessment_1999}.%

\section{Simulation study} \label{sec:sim}
The purpose of this section is three-fold. First, we illustrate how the proposed estimation framework scales well to large sample sizes, commonly seen in the credit context. Second, by using simulation setups identical to those presented by \cite{xie_promotion_2021}, we show the computational advantages of estimating the model through end-to-end trained DNN architecture versus iteratively optimizing the DNN in the maximization step of the EM algorithm. Finally, we analyze how the orthogonalization procedure can recover the structured linear predictor without compromising performance compared to the setting without orthogonalization (unrestricted).

\paragraph*{Simulation Design} We study three sample sizes $N$: 50,000, 100,000, and 150,000 subjects. The sample sizes from the works presented in Table \ref{tab:refs_cure} have, on average, $\sim 30,000$ subjects, with a maximum $N$ of 80,641. Therefore, we consider 50,000 as a relevant starting sample size in this context and scale it to 150,000, which is roughly the size of our dataset (and the largest we are aware of). One can argue that the cure models used for credit applications so far and commonly estimated via the EM algorithm, could not be scaled due to computational burdens. Our framework does not have those constraints. 

We evaluate four simulated scenarios: three presented in \cite{xie_promotion_2021} and a fourth in which we added a linear component to study the orthogonalization feature. All scenarios are described in detail in Appendix \ref{app:1}.

\paragraph*{Summary of Results}
Table \ref{tab:sim_res} shows the comparison between the EM implementation (EM-PTCM) versus the Deep-PTCM for each combination of sample size ($N$) and scenario. The column \textbf{Time} is the average time in minutes needed to estimate the model for the corresponding setting. Moreover, the columns $\Delta S$, $\Delta  S_p$, and $\Delta \eta$ show the mean square difference between the true and estimated quantities $S$, $S_p$, and $\eta$, respectively. That is e.g.\ $\Delta S=\frac{1}{R\cdot N}\sum_{r=1}^{R}\sum_{i=1}^N (\hat{S}^{(r)}(t_i;\mathbf{x}_i)-S^{(r)}(t_i;\mathbf{x}_i))^2$, where $\hat S^{(r)}(\cdot)$ and $S^{(r)}(\cdot)$ are, respectively, the estimated and the true survival function for replication $r$. The other cases follow analogously. These metrics are evaluated on 100 ($R=100$) holdout datasets with the same data generation process but different random seeds.

\begin{table*}[ht]
\renewcommand{\arraystretch}{1.3}
\caption{Simulation results for 100 independent replications. Time is the average time in minutes needed for training.}
\label{tab:sim_res}
\centering
\begin{tabular}{lc llll llll}
\toprule
 &  & \multicolumn{4}{S}{\textbf{EM-PTCM}} & \multicolumn{4}{S}{\textbf{Deep-PTCM}} \\
$N$ & \textbf{Scenario} & \textbf{Time} & $\Delta S$ & $\Delta S_p$ & $\Delta \eta$ & \textbf{Time} & $\Delta S$ & $\Delta S_p$ & $\Delta \eta$ \\
\midrule
\multirow[c]{3}{*}{50,000} & 1 & 32.4 & \textbf{0.0001} & 0.0002 & \textbf{0.0113} & \textbf{0.5} & 0.0029 & \textbf{0.0001} & 0.0167 \\
 & 2 & 18.5 & 0.0007 & 0.0003 & 0.0166 & \textbf{0.3} & \textbf{0.0005} & \textbf{0.0003} & \textbf{0.0065} \\
 & 3 & 26.1 & \textbf{0.0002} & 0.0052 & 0.1037 & \textbf{1.0} & 0.0022 & \textbf{0.0030} & \textbf{0.0886} \\
\multirow[c]{3}{*}{100,000} & 1 & 106.9 & \textbf{0.0001} & 0.0002 & \textbf{0.0096} & \textbf{0.7} & 0.0016 & \textbf{0.0001} & 0.0101 \\
 & 2 & 58.4 & \textbf{0.0006} & 0.0003 & 0.0136 & \textbf{0.7} & 0.0012 & \textbf{0.0002} & \textbf{0.0075} \\
 & 3 & 61.4 & 0.0002 & 0.0051 & 0.1028 & \textbf{1.2} & \textbf{0.0001} & \textbf{0.0025} & \textbf{0.0673} \\
\multirow[c]{3}{*}{150,000} & 1 & 187.7 & \textbf{0.0001} & 0.0002 & \textbf{0.0081} & \textbf{0.9} & 0.0026 & \textbf{0.0001} & 0.0196 \\
 & 2 & 100.8 & 0.0006 & 0.0003 & 0.0125 & \textbf{0.9} & \textbf{0.0001} & \textbf{0.0001} & \textbf{0.0025} \\
 & 3 & 89.9 & \textbf{0.0004} & 0.0060 & 0.1187 & \textbf{1.3} & 0.0006 & \textbf{0.0025} & \textbf{0.0692} \\
\bottomrule
\end{tabular}
\end{table*}
The minimum between both approaches is shown in bold. We observe that the mean square differences, for $S_p$ and $\eta$, are generally lower for Deep-PTCM than for EM-PTCM. In the case of $S$, this difference is not so clear. Nevertheless, since both implementations are meant to estimate the same model, it is not surprising that these results are indeed comparable. The great advantage, however, is that the Deep-PTCM is significantly faster than the EM implementation (more than 100 times for some cases) without compromising accuracy. 

\begin{figure}[ht]
    \centering
    \includegraphics[width=0.7\textwidth]{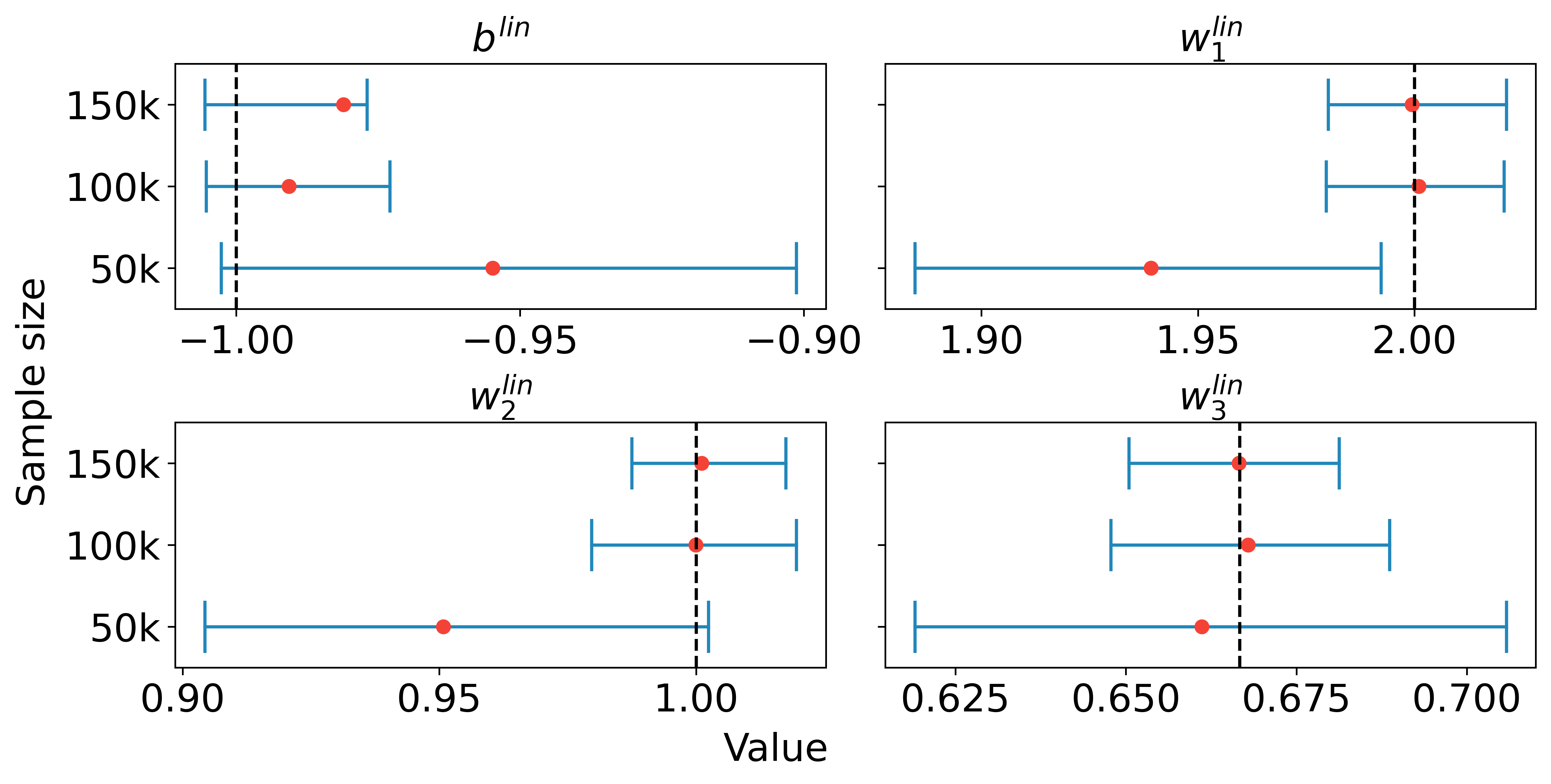}
    \caption{Linear coefficients estimated by Deep-PTCM with orthogonalization (Deep-PTCM-Ort).}
    \label{fig:sim_coeffs}
\end{figure}

To analyze orthogonalization, we create a fourth setting with $\eta=\eta^{lin}+\eta^{non}$ in which $\eta^{lin} = 
b^{lin}+w^{lin}_1x_1+w^{lin}_2x_2+w^{lin}_3x_3$ and $\eta^{non}$ follows similarly to the one defined in \textit{Scenario 2}. Table \ref{tab:sim_res_ort} summarizes the results from 100 independent replications comparing the Deep-PTCM and its version with orthogonalization. We note that, in general, the performance of both models is similar concerning the mean square differences. Moreover, Figure \ref{fig:sim_coeffs} depicts the 2.5-97.5\% range of the estimations of $b^{lin}$, $w^{lin}_1$, $w^{lin}_2$ and $w^{lin}_3$ across the 100 replications. We observe a suitable recovery of the true parameter values (dashed vertical lines), especially when increasing the sample size.

\begin{table*}[ht]
    \renewcommand{\arraystretch}{1.3}
    \caption{Simulation results for 100 independent replications without (Deep-PTCM) and with (Deep-PTCM-Ort) orthogonalization.}
    \label{tab:sim_res_ort}
    \centering
    \begin{tabular}{l llll llll}
        \toprule
        & \multicolumn{4}{c}{\textbf{Deep-PTCM}} & \multicolumn{4}{c}{\textbf{Deep-PTCM-Ort}} \\
       \textbf{N} & \textbf{Time} & $\Delta S$ & $\Delta S_p$ & $\Delta \eta$ & \textbf{Time} & $\Delta S$ & $\Delta S_p$ & $\Delta \eta$ \\
       \midrule
    50,000 & \textbf{0.1} & \textbf{0.0001} & \textbf{0.0006} & \textbf{0.0263} & 0.2 & 0.0023 & 0.0029 & 0.0985 \\
    100,000 & \textbf{0.2} & 0.0001 & 0.0001 & 0.0133 & 1.6 & \textbf{0.0000} & \textbf{0.0000} & \textbf{0.0005} \\
    150,000 & \textbf{0.2} & \textbf{0.0000} & 0.0001 & 0.0122 & 1.5 & 0.0021 & \textbf{0.0000} & \textbf{0.0106} \\
       \bottomrule
    \end{tabular}
\end{table*}

\section{Application} \label{sec:app}
\subsection{Data}
We analyze the publicly available single-family loan-level dataset from Freddie Mac\footnote{\url{http://www.freddiemac.com/research/datasets/sf_loanlevel_dataset.page}}. The dataset contains loan-level origination and monthly performance for fixed-rate US mortgages and is periodically updated. The event of interest is the credit default, defined as the moment the loan is past due in 90 or more days. The training set contains 149,561 loans granted between 2009 and 2011. The test set includes 49,888 loans granted in 2013. The monitoring periods for both sets date from loan origination to December 2021. The data include eleven categorical and eight numerical variables. Tables \ref{tab:cat_vars} and \ref{tab:num_vars} in the Appendix describe the categorical and the numerical variables, respectively.

Some categorical variables present high cardinality, which can be challenging from an estimation perspective (poor generalization and high resource usage \cite{carneiro2022high}). Common practice is to either drop these variables, thus discarding valuable information or to transform the attributes into numerical representations, such as target encoding \cite{micci2001preprocessing}. To compare different preprocessing practices for these variables when estimating the standard PTCM, we employ target encoding, one-hot encoding, and principal component analysis (PCA) for dimensionality reduction. On the other hand, when estimating the deep version, we only use one-hot encoding, arguing that the DNN should be able to generalize well without further preprocessing steps. The dimension of the feature space after encoding is 921. We normalized all numerical variables to make the training procedure more efficient.
\begin{figure}[ht]
    \centering
    \includegraphics[width=0.7\textwidth]{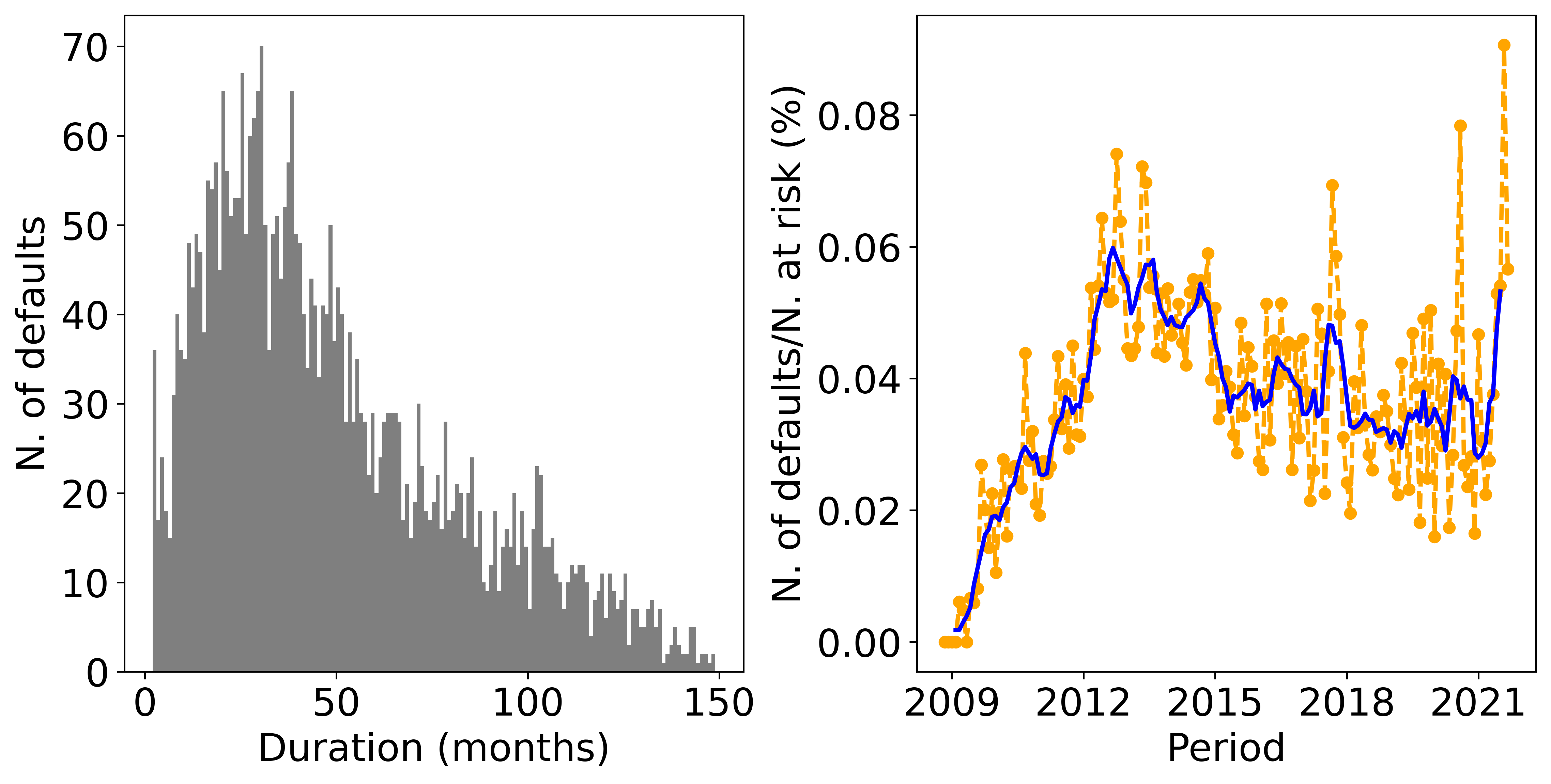}
    \caption{Single-family loan-level dataset from Freddie Mac. Left:  distribution of default events versus  duration. Right: ratio between the number of default events and  borrowers at risk over calendar time. The solid blue line is the moving average for a six-month window.}
    \label{fig:df_rt}
\end{figure}

Figure \ref{fig:df_rt} illustrates the distribution of default events as a function of duration (left) and the number of defaults with respect to the borrowers at risk over the calendar time (right). The solid blue line corresponds to the moving average for a six-month window.

\subsection{Network Architecture and Training} \label{sec:architecture}
For the available data, we use an FCFN in the \texttt{DNN} block (cf.~Figure \ref{fig:nnet}). This network architecture is often employed in tabular data settings and can be effective for learning complex relationships within features. Although we use this type of network to predict the time to default in a mortgage portfolio, we emphasize that since the Deep-PTCM implementation uses the TensorFlow framework, any architecture available there can be adapted to the needs of other applications.

To train the network, the architecture of the FCFN needs to be tuned to achieve high prediction accuracy. This includes defining the number of layers, the number of units for each layer, the activation functions, which optimization algorithm to use, etc.\ These hyperparameters are not learned during backpropagation and must be set manually. Hyperparameters can significantly impact the performance of the model, and several tuning strategies have been proposed \cite{he2021automl}. We use the random search strategy commonly employed in DL \cite{bergstra2012random}. We then chose the combination of hyperparameters that accomplished the minimum average loss in three execution runs per trial with independent random initializations. Running each trial multiple times means avoiding making the final decision strictly dependent on the initial random values.%

In addition, to prevent overfitting when training each trial, we use early stopping. We set a maximum number of epochs and a \say{patience} parameter, which is the number of epochs to wait before deciding to stop the training process. During training, we monitor the model performance on a hold-out or validation set not used in the optimization. If the model performance on the validation set does not improve for a certain number of epochs (the patience parameter), then the training is stopped.

After hyperparameter tuning, the resulting network architecture for the \texttt{DNN} block of the Deep-PTCM has three layers (two hidden plus the output layer). The first two hidden layers have 512 units, {ReLU} activation functions, and a {dropout} rate of 0.2. The output layer, representing the predictor $\eta$, has one unit and a linear activation function. The best optimization algorithm was found to be stochastic gradient descent (SGD; see \cite{zhang2021dive}, Chap.~12.4) with a learning rate schedule that follows the inverse time decay with an initial rate of 0.01, a decay rate of 0.75, and 100 decay steps. The final model is re-trained using the entire training set (see Table \ref{tab:tuning} in the Appendix for further details on the search space of the hyperparameters).

\subsection{Results} \label{sec:results}
\begin{figure}[ht]
    \centering
    \includegraphics[width=0.7\textwidth]{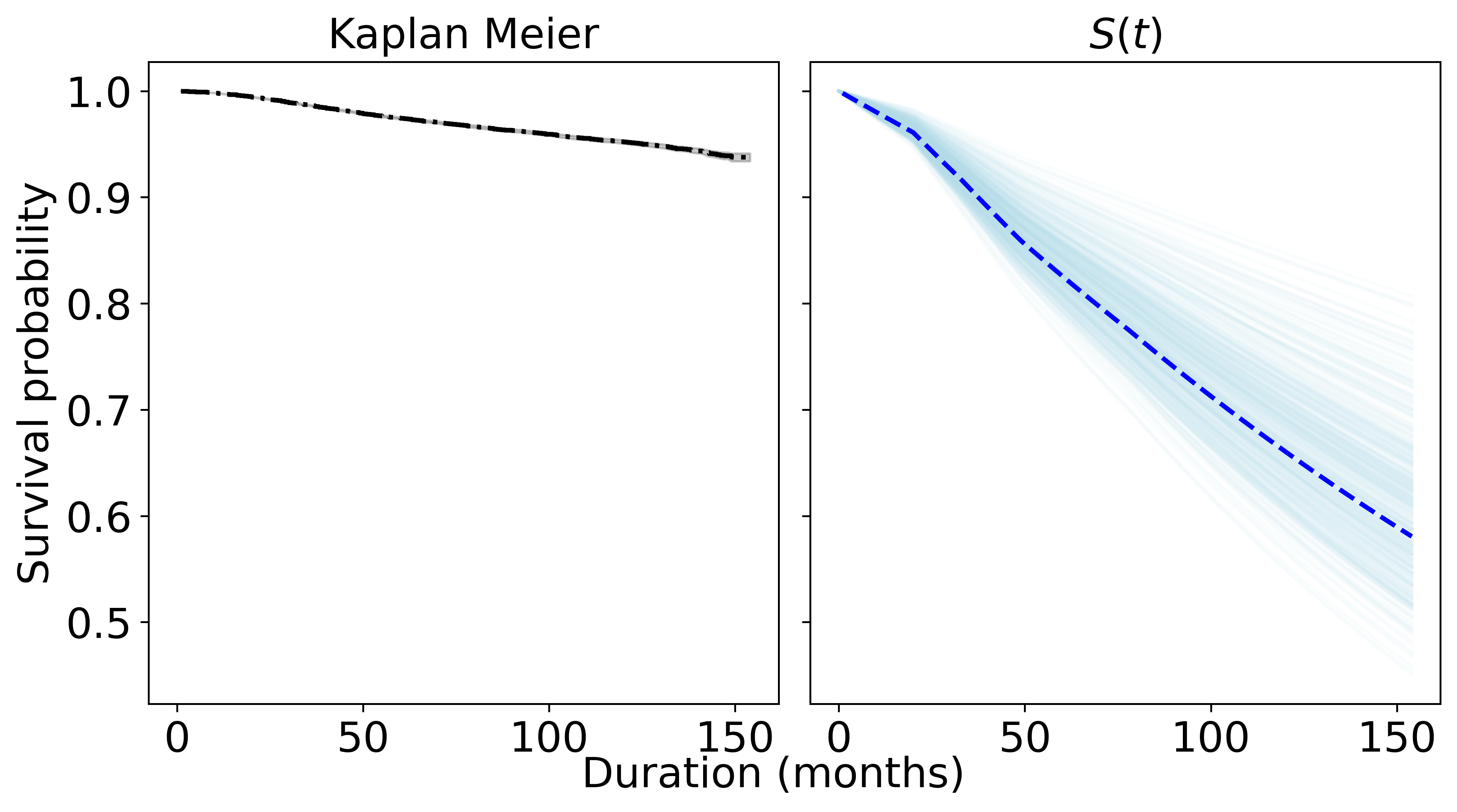}
    \caption{Left: Kaplan-Meier curve. Right:   survival function of the risk factors $S(t)$  (dashed). The blue-shaded curves are those obtained by 500 bootstrap samples with replacement.}
    \label{fig:km_S1}
\end{figure}

Figure \ref{fig:km_S1} illustrates the Kaplan-Meier curve for the whole population (left) and the estimated survival function $S(t) = 1-F(t)$ of the risk factors (right). The transparent curves represent 500 estimations of the survival function based on resampling with replacement. The interpretation of $S(t)$ is sometimes mistakenly considered as the survival function of non-cured subjects (e.g.\ \cite{xie_promotion_2021}). But since $F(t)$ is the CDF of the risk factors, and the time to the event is when the first one is triggered (cf.~Section \ref{subsec:ptcm}), $S(t)$ represents an upper bound of the survival function of the susceptible individuals \cite{peng2012extended}. Therefore, since the Kaplan-Meier estimator does not control for cured and non-cured subjects, it calculates, for instance, that the probability of default, or \say{not surviving}, would be $\sim 5\%$ after ten years of payments. However, the Deep-PTCM estimates that if the subject belonged to the susceptible population, the probability of default would not be lower than $\sim 35\%$. 

Studying the predictive power of credit scoring models is relevant from the perspective of credit risk management \cite{thomas2017credit}. In particular, we are interested in how the Deep-PTCM performs compared to the traditional PTCM and what gains the Deep-PTCM offers. To this end, we consider five models. The first three correspond to different versions of the PTCM with linear effects in the predictor, where the preprocessing technique of the features gives the distinction. The first one employs one-hot encoding (\textit{PTCM}), the second target encoding (\textit{ENC-PTCM}), and the third PCA for dimensionality reduction (\textit{PCA-PTCM}). The purpose is to apply, on the one hand, the standard practices when modeling in the presence of high-cardinality categorical variables and, on the other, to make the comparison to the deep version more comprehensive. The other two models, \textit{Deep-PTCM} and \textit{Deep-PTCM-Ort}, correspond to the deep approach where the difference is that the latter applies orthogonalization.

Table \ref{tab:perf_comp} depicts the results obtained on the test dataset for the performance metrics described in Section \ref{sec:perf}. The numbers in parenthesis are the standard deviations obtained from 100 bootstrap samples of the same size as the original data. We notice that among the three PTCMs, the best discrimination, as measured by $\text{AUC}_{\text{cure}}$, is obtained by the version with one-hot encoding (\textit{PTCM}). In terms of calibration, as measured by IBS, the \textit{PCA-PTCM} showed the minimum among the three, but the difference is not significant. Moreover, compared to the deep versions, neither of the three PTCMs with linear predictors performed better in discrimination. Between the two deep versions, we note orthogonalization does not improve predictive performance for this case study. However, interpretability gains are, of course, always present. The best results for discrimination and calibration are accomplished by \textit{Deep-PTCM}, showing an $\text{AUC}_{\text{cure}}$ of 0.88, compared to 0.85 from \textit{PTCM}, and an IBS of 0.022, compared to 0.023 from \textit{PCA-PTCM}. 

\begin{table}[ht]
    \renewcommand{\arraystretch}{1.3}
    \caption{$\text{AUC}_{\text{cure}}$ and IBS results evaluated in the test set.}%
    \label{tab:perf_comp}
        \centering
		\begin{tabularx}{0.7\textwidth}{lXXXXX}
			\toprule
			{} &                 PTCM &            ENC-PTCM &             PCA-PTCM &            Deep-PTCM & Deep-PTCM-Ort \\
			\midrule
			\multirow[c]{2}{*}{$\text{AUC}_{\text{cure}}$} &  0.85305 &  0.82245 &  0.84524 &  \textbf{0.88301} &  0.8628 \\
			& (0.00094) & (0.00077) &  (0.00079) &  (0.00073) &  (0.00074) \\
			\multirow[c]{2}{*}{IBS} &  0.02317 &  0.02351 &  0.02299 &   \textbf{0.02231} &  0.02362 \\
			&  (0.00062) &  (0.00063) &  (0.00061) &   (0.00059) &  (0.00063) \\
			\bottomrule
		\end{tabularx}
\end{table}

\begin{figure}
    \centering
    \includegraphics[width=0.7\textwidth]{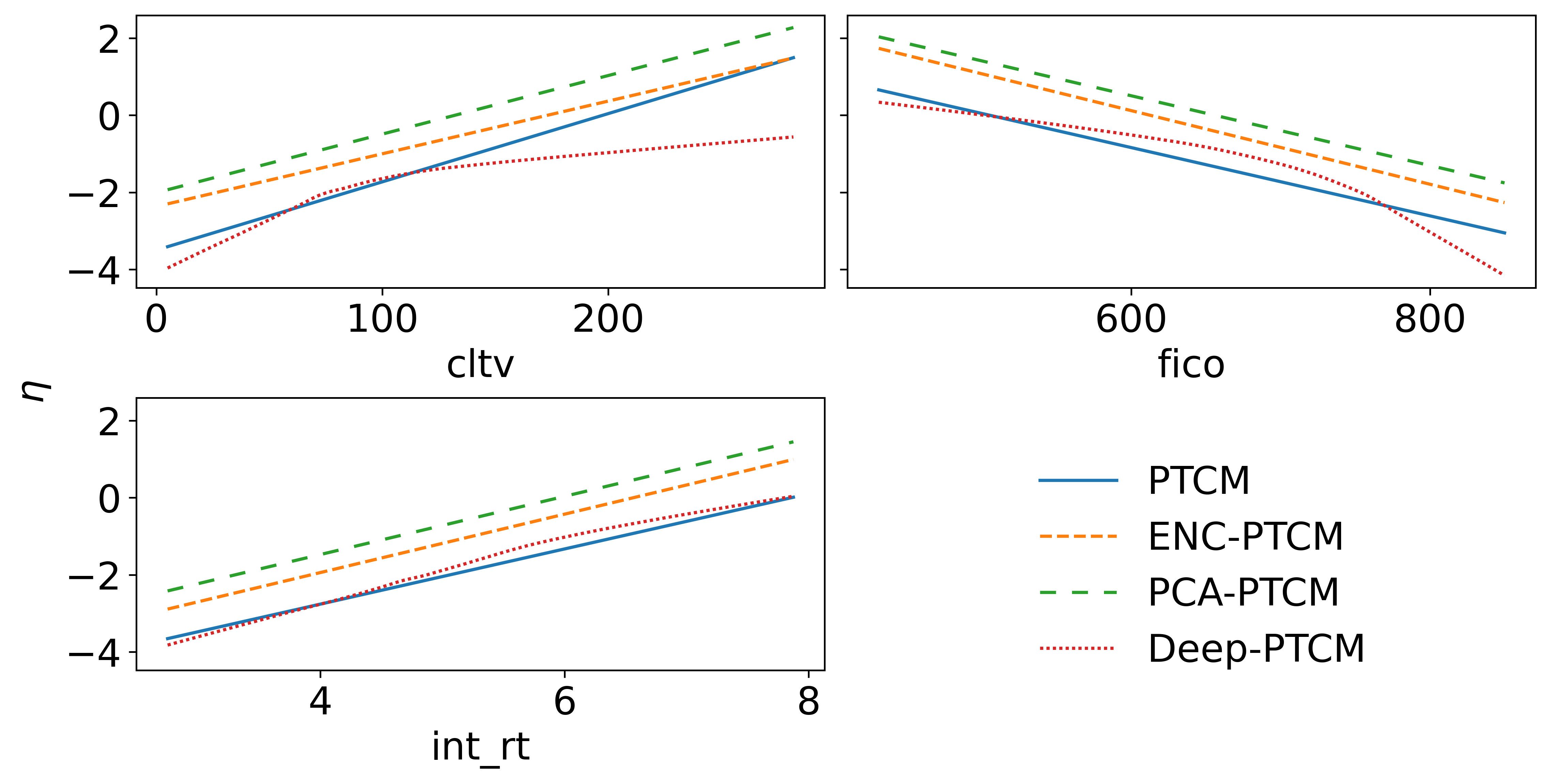}
    \caption{Comparison of the effect of numerical covariates on the predictor $\eta$ for four cure models.}
    \label{fig:cov_eff}
\end{figure}

To illustrate how the best-performing model, \textit{Deep-PTCM}, measures the effect of the numeric covariates in $\eta$ and how it compares to the linear ones, we emulate fictional borrowers with all the covariates centered in their based values but the one in question, for which we recreate a grid within its range. Results for three numeric covariates are presented in Figure \ref{fig:cov_eff}. One can first notice that the deep version measures non-linear relationships between some covariates and the predictor. Two remarkable examples are the combined loan-to-value ratio \textit{cltv} and the borrower's external credit score \textit{fico}. Both covariates have shown conforming signs in the credit risk literature when linearity is assumed \cite{wang2020reducing, medina2022joint}. Greater values of \textit{cltv} are associated with a greater risk of default, and greater \textit{fico} values are associated with lower risk. We show the same trend but in a non-linear way.

In the \textit{cltv} case, we observe that the \textit{Deep-PTCM}, like the \textit{PTCM}, reckon similar risk increments between 50 and 120. However, for values lower than 50 or above 120, the risk estimated by the deep version is lower. For \textit{fico}, we observe that the effect between 550 and 750 calculated by \textit{Deep-PTCM} is more significant than the one shown by \textit{PTCM}. Yet, if the score assigned by the credit bureau is higher than $\sim 750$ (good creditworthiness), the risk measured by the deep version starts to go down comparatively. 

In addition, we note that there are covariates, such as the interest rate \textit{int\_rt}, where both the \textit{PTCM} and the \textit{Deep-PTCM}, estimate a linear relationship, despite the fact the last one is not restricted to do so. The effects of the other numerical covariates are in the Appendix.

\begin{figure}[ht]
    \centering
    \includegraphics[width=0.7\textwidth]{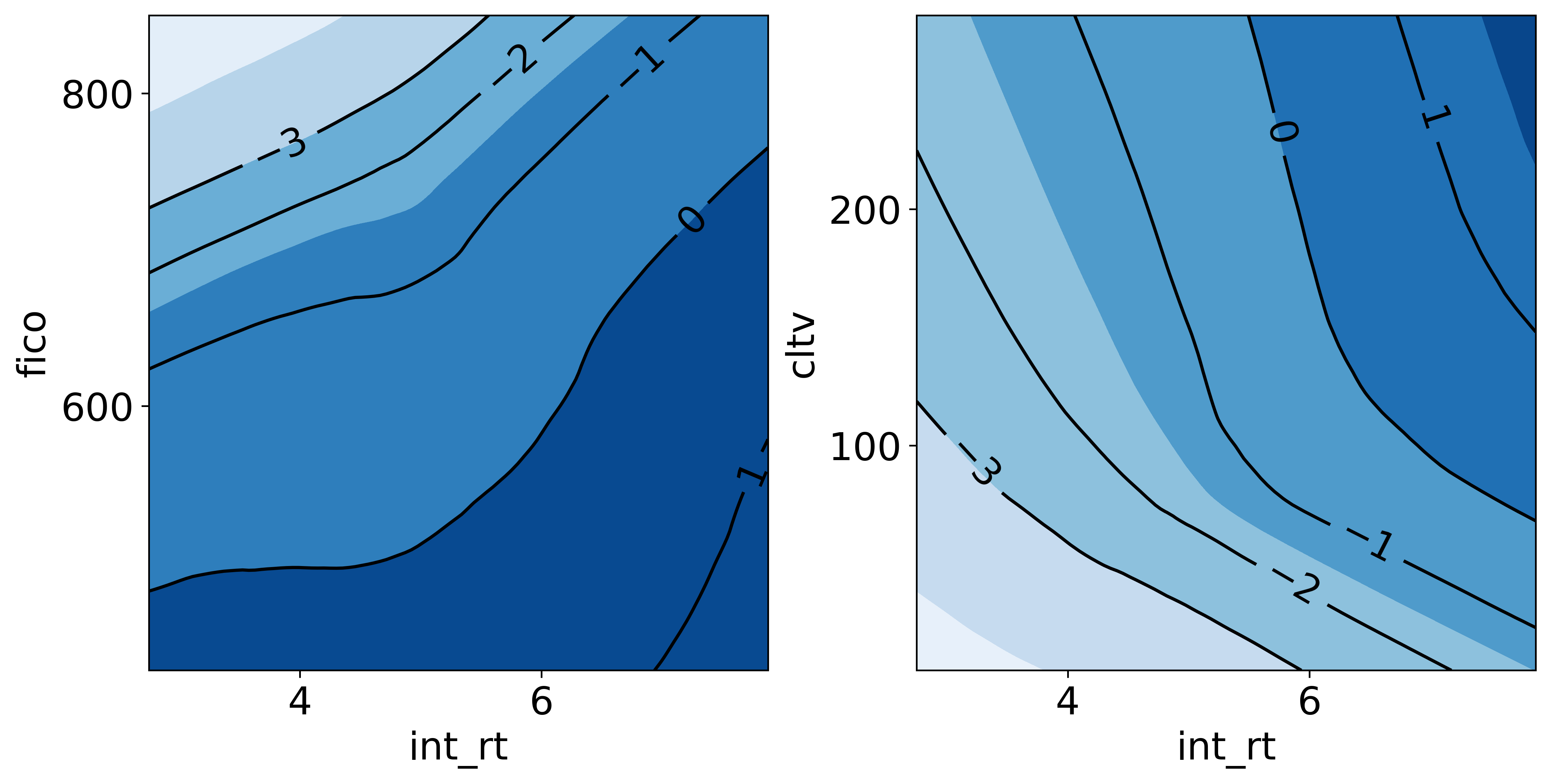}
    \caption{Bivariate interactions slices of the predictor $\eta$ for covariates \textit{int\_rt}-\textit{fico} (left) and \textit{int\_rt}-\textit{cltv} (right).}
    \label{fig:cov_inter}
\end{figure}

However, the Deep-PTCM can reveal not only the non-linearities of single covariate effects but also potential interactions. To illustrate this, Figure \ref{fig:cov_inter} visualizes a slice of the bivariate interaction of the pair \textit{int\_rt}-\textit{fico} and \textit{int\_rt}-\textit{cltv}. We observe that the effect of \textit{int\_rt} for values of \textit{fico} less than 600 does not change substantially. Similarly, we see that for loans with interest rates close to 6\%, the effect of \textit{cltv} is maintained for values greater than 80. The traditional PTCM cannot provide this information.

Overall, we conclude that the Deep-PTCM is able to recover the simpler embedded PTCM without requiring this (often too restrictive) assumption to be made in advance.

The estimates for the linear effects further support this conclusion. For the variables \textit{cltv}, \textit{fico}, and \textit{int\_rt}, these are 0.30/0.26, -0.41/-0.52, and 0.39/0.38 for the \textit{PTCM} and \textit{Deep-PTCM-Ort} models, respectively (see the Appendix for more details). However, the increased flexibility of \textit{Deep-PTCM-Ort} comes at the cost of having more unknown parameters and, therefore, greater parameter uncertainty. Nevertheless, given the slightly better prediction performance of \textit{Deep-PTCM} and our simulations, which demonstrate that uncertainty can be significantly reduced with more data, we believe our approach makes a valuable and innovative contribution to future large-scale credit risk applications.%

\section{Discussion} \label{sec:concl}
Survival models assume that all individuals are, sooner or later, prone to the event of interest. However, there are applications, such as mortgage default prediction, where it is noticeable that some are not susceptible to the event. Under these circumstances, cure rate models are preferable. The literature on credit risk modeling with cure fractions mostly considers the MCM approach (cf.~Table \ref{tab:refs_cure}). However, another class of models, the PTCM, has not received the same level of attention in this context.

We propose a reformulation of the PTCM, the Deep-PTCM, that simultaneously estimates the covariate effects and the parameters associated with the survival distribution in an end-to-end DNN. This allows us, on the one hand, to account for complex and often more realistic non-linear relationships between covariates and survival and, on the other, to have a computationally efficient approach that scales well to large datasets, such as the ones seen in credit risk applications. Moreover, interpretability can be crucial in some fields when choosing a model. While DL offers flexibility and superior predictive power, it has its critics regarding explainability. Our approach tries to leverage the advantages of DL with an orthogonalization method that facilitates interpretability. We aim to provide an accurate and transparent framework, allowing decision-makers to make informed decisions without making restrictive linearity assumptions a priori. 

Via simulations, we demonstrate the scalability of our method compared to an existing one based on the EM algorithm, reducing, for some cases, the average training time to the one-hundredth part. In addition, we show that the Deep-PTCM can significantly improve discrimination and calibration metrics compared to the standard PTCM when predicting the time to default in a large US mortgage portfolio. Finally, we further explore how DNN flexibility accounts for the effects of the covariates on the predictor, observing, first, its ability to correctly detect present deviations from linearity in the predictor as assumed by the classic PTCM and, second, to recover it if the evidence supports it.

Despite concentrating on one specific use case, credit risk modeling, in this paper, we emphasize that the statistical and computational properties of the Deep-PTCM ensure broader applicability in modeling time-to-event data in other fields. To facilitate this, we provide a Python package, \texttt{deepcure} that accompanies the paper and is available on \href{https://github.com/vhmedina/deepcure}{GitHub}.  

Ultimately, we envision high-potential paths of research for the Deep-PTCM. In particular, we can extend the framework to include covariate effects on both the predictor $\eta$ and the CDF $F$ \cite{tsodikov2002semi}. One way is to create a second \texttt{DNN} block that takes as input, in addition to $\mathbf{t}$ and $\bm{\delta}$, a covariate matrix (can be the same as $\mathbf{X}$ or not), and has as outputs $F$ and $f$. Together with $\eta$ from the first \texttt{DNN} block, these would be the new inputs to the \texttt{Endpoint Layer}. The appeal of this setup is that we can analyze short- and long-term effects without strong parametric assumptions in $F$ by including covariates and modeling them via a \texttt{DNN} block.%

\section*{Acknowledgments}
{This work has been partly supported by the \href{https://rc-trust.ai}{Research Center Trustworthy Data Science and Security}, one of the Research Alliance centers within the \href{https://uaruhr.de}{UA Ruhr}. Nadja Klein acknowledges funding by the Deutsche Forschungsgemeinschaft (DFG, German Research Foundation) through the Emmy Noether grant KL 3037/1-1.%
}

\bibliographystyle{abbrv}
\bibliography{refs} 

\appendix

\section{Simulation Scenarios} \label{app:1}

We consider the following scenarios in Section \ref{sec:sim}.
\paragraph*{Scenario 1} $p=1$ with $\theta(x) = 0.15\exp\{3.5 \cdot 10^3x^2(1-x)^8+2.2\cdot 10^4x^8(1-x)^3\}$
\paragraph*{Scenario 2} $p=3$ with $\theta(\mathbf{x})=\exp\{-0.8x_{1}^2+4x_{2}^3-0.75\cos(x_{3})\}$
\paragraph*{Scenario 3} $p=10$ with $\theta(\mathbf{x})= 0.4\big[0.05\big(x_{1}^2+\tanh(x_{2})-x_{3}\cdot x_{4}(4-0.0005x_{3}\cdot x_{4})^2+\log(|x_{1}+x_{5}|^{20})\big)\big]+ 0.05\big[x_{6}^2+\tanh(x_{7})-x_{8}\cdot x_{9}(4-0.0005x_{8}\cdot x_{9})^2+\log(|x_{6}+x_{10}|^{20})\big]$
\paragraph*{Scenario 4} $p=3$ with $\eta(\mathbf{x}) = \eta^{lin}(\mathbf{x})+\eta^{non}(\mathbf{x})$, where $\eta^{lin}(\mathbf{x})= -1+\frac{2}{1}x_1+\frac{2}{2}x_2+\frac{2}{3}x_3$ (following \cite{rugamer2022semi}) and $\eta^{non}(\mathbf{x})=-0.8x_{1}^2+4x_{2}^3-0.75\cos(x_{3})$, as in \textit{Scenario 3}. $\eta^{non}(\mathbf{x})$ is projected to the orthogonal complement of $\eta^{lin}(\mathbf{x})$.

The covariates for Scenarios 1 and 2 are randomly sampled from a uniform distribution between 0 and 1. For Scenario 3, $x_1$ to $x_5$ follow a multivariate normal distribution with covariance matrix
\begin{displaymath}
    \Sigma = \begin{pmatrix}
    1 & 0.8 & 0.5 & 0.2 & 0 \\
    0.8 & 1 & 0.2 & 0.6 & 0 \\
    0.5 & 0.2 & 1 & 0.3 & 0 \\
    0.2 & 0.6 & 0.3 & 1 & 0 \\
    0 & 0 & 0 & 0 & 1 
    \end{pmatrix}.
\end{displaymath}

The other five covariates, $x_6$ to $x_{10}$, are independent and follow standard normal distributions. Furthermore, the failure times for all scenarios are generated from an exponential distribution with a rate parameter of 1. However, as described in Section \ref{subsec:deepptcm} and following \cite{chen_maximum_2001, xie_promotion_2021}, the specification for estimating $F(t)$ is the piecewise exponential.

\section{Description and Statistics of the Data}\label{app:data}

\begin{table}[H]
    \caption{Categorical variables in the single-family loan-level dataset from Freddie Mac.}
\label{tab:cat_vars}
\centering
\begin{tabularx}{\textwidth}{l S X}
\toprule
\textbf{Variable} & \textbf{Unique Values} & \textbf{Description} \\
\midrule
channel & 3 & Whether a Broker/Correspondent/Retail originates the mortgage loan \\
cnt\_borr & 2 & Number of borrowers on the loan (1, 2, or not available) \\
dti & 7 & Debt-to-income ratio of the borrower(s) (discretized into 7 categories) \\
flag\_fthb & 2 & Indicates whether the borrower(s) is(are) first-time homebuyers (yes/no/not applicable) \\
flag\_sc & 2 & If the mortgage exceeds conforming loan limits (yes/no) \\
loan\_purpose & 3 & Indicates whether the loan is a purchase/cash out refinance/no cash out refinance mortgage \\
occpy\_sts & 3 & Denotes whether the loan is owner occupied/second home/investment property \\
prop\_type & 5 & Denotes whether the property type secured by the mortgage is a condo/planned unit development/manufactured housing/single-family/cooperative share \\
rel\_ref\_ind & 2 & Indicates if the loan is part of a relief refinance program (yes/no) \\
servicer\_name & 24 & Entity acting in its capacity as the servicer of mortgages to Freddie Mac (24 servicers) \\
zipcode & 860 & First 3 digits of the zipcode of the property \\
\bottomrule
\end{tabularx}

\end{table}

\begin{table}[H]
    \caption{Numerical variables in the single-family loan-level dataset from Freddie Mac.}
\label{tab:num_vars}
\centering
\begin{tabularx}{\textwidth}{l S S S S X}
\toprule
\textbf{Variable} & \textbf{Mean} & \textbf{Std} & \textbf{Min} & \textbf{Max} & \textbf{Description} \\
\midrule
cltv & 70.2 & 20.0 & 5.0 & 282.0 & Combined loan-to-value ratio obtained by dividing the original loan amount plus any additional mortgage debt by the appraised value of the property \\
cnt\_units & 1.0 & 0.2 & 1.0 & 4.0 & Denote the number of units in the mortgaged property \\
fico & 760.1 & 44.3 & 431.0 & 850.0 & Number summarizing the borrower's external credit score \\
int\_rt & 4.7 & 0.5 & 2.8 & 7.9 & The original interest rate of the loan \\
ltv & 68.0 & 19.0 & 5.0 & 228.0 & Original loan-to-value ratio \\
mi\_pct & 1.8 & 6.5 & 0.0 & 40.0 & The percentage of loss coverage on the loan in case of default \\
orig\_loan\_term & 303.4 & 82.2 & 60.0 & 360.0 & Number of scheduled payments of the mortgage \\
orig\_upb & 207.5 & 116.2 & 8.0 & 1000.0 & Original loan amount (in thousands of dollars) \\
\bottomrule
\end{tabularx}

\end{table}

\section{Search Space of the Hyperparameter Tuning Process} \label{app:tuning}

\begin{table}[H]
    \caption{Search space options for the hyperparameter tuning process.}
\label{tab:tuning}
\centering
\begin{tabular}{ll}
\toprule
\textbf{Feature} & \textbf{Options} \\
\midrule
Layers & $\{1,2,3\}$ \\
Batch normalization & \{Yes,No\} \\
Units & $\{64, 128, 192, 256, 320, 384, 448, 512\}$ \\
Activation function & $\{\text{tanh},\text{ELU},\text{ReLU},\text{sigmoid}\}$ \\
Dropout rate & $\{0.20, 0.35, 0.50\}$ \\
Optimizer & $\{\text{Adam},\text{RMSprop},\text{SGD}\}$ \\
Decay steps & $\{10,100,1000\}$ \\
Decay rate & $\{0.50, 0.75, 0.90\}$\\
Learning rate schedules & \{Exponential, Inverse time, Cosine\}\\
\bottomrule
\end{tabular}

\end{table}

\section{Comparison of the effect of numerical covariates on the predictor}

\begin{figure}[H]
    \centering
    \includegraphics[width=0.7\textwidth]{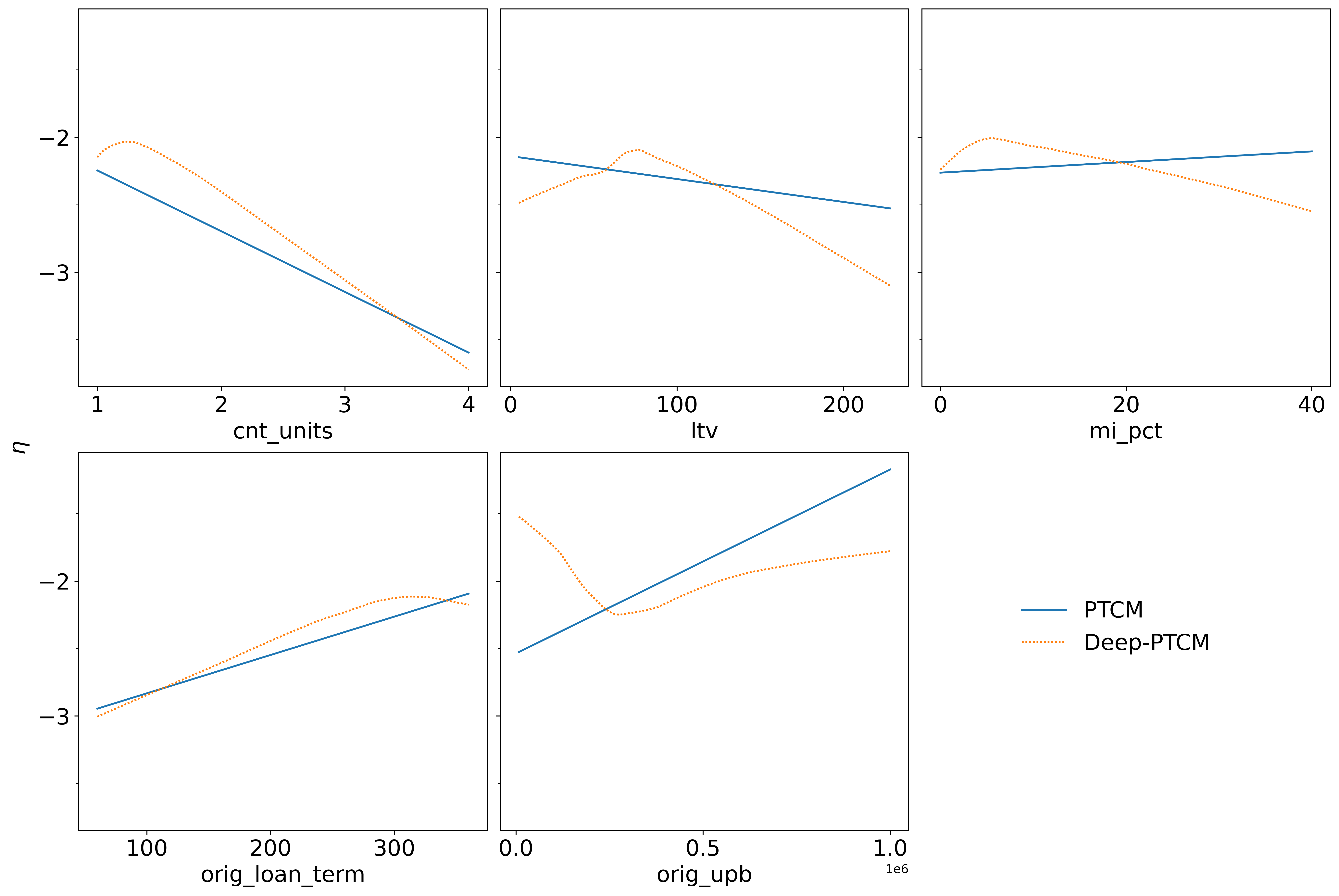}
    \caption{Comparison of the effect of \textit{cnt\_units},\textit{ltv}, \textit{mi\_pct}, \textit{orig\_loan\_term}, \textit{orig\_upb} on the predictor $\eta$ for PTCM and Deep-PTCM.}
    \label{fig:cov_eff2}
\end{figure}

\section{Linear coefficient estimates of numerical covariates}

\begin{figure}[H]
    \centering
    \includegraphics[width=0.7\textwidth]{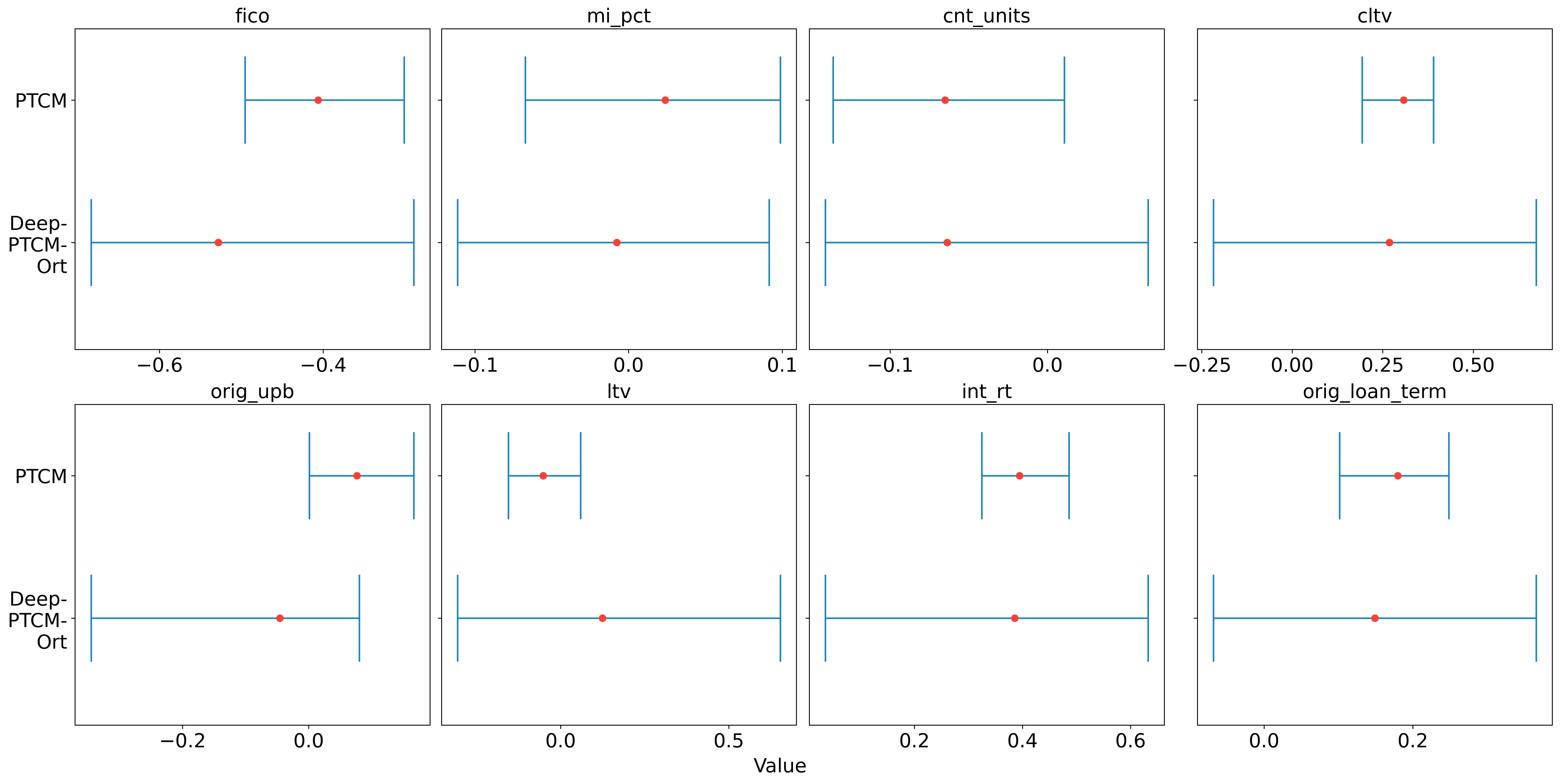}
    \caption{Comparison of the linear coefficients estimates from PTCM and Deep-PTCM-Ort.}
    \label{fig:lin_coeff_all}
\end{figure}

\end{document}